\documentclass[journal]{IEEEtran}
\usepackage{hyperref}
\usepackage{cite}
\usepackage[shortlabels]{enumitem}
\usepackage{graphicx}
\usepackage{multirow}
\usepackage{amsmath}
\usepackage{makecell}
\usepackage{footnote}
\makesavenoteenv{tabular}
\usepackage{amssymb}
\begin{document}

\title{A Federated Learning Framework for Smart Grids: \\ Securing Power Traces in Collaborative Learning}

\author{\IEEEauthorblockN{Haizhou Liu, Xuan Zhang*, Xinwei Shen*,~\IEEEmembership{Senior Member,~IEEE}, and
Hongbin Sun,~\IEEEmembership{Fellow,~IEEE}}

\thanks{All authors are affiliated with Tsinghua-Berkeley Shenzhen Institute, Tsinghua Shenzhen International Graduate School, Tsinghua University, 518055, Shenzhen, China. Hongbin Sun is also affiliated with the Department of Electrical Engineering, Tsinghua University, 100084, Beijing, China. (Corresponding Author: Xuan Zhang, \href{mailto:xuanzhang@sz.tsinghua.edu.cn}{xuanzhang@sz.tsinghua.edu.cn}; Xinwei Shen, \href{mailto:sxw.tbsi@sz.tsinghua.edu.cn}{sxw.tbsi@sz.tsinghua.edu.cn}.)\\ \null \quad This paper is currently being revised and submitted for peer-review in the name of ``Privacy-Preserving Power Consumption Prediction Based on Federated Learning with Cross-Entity Data".} }


\IEEEtitleabstractindextext{%
\begin{abstract}
With the deployment of smart sensors and advancements in communication technologies, big data analytics have become vastly popular in the smart grid domain, informing stakeholders of the best power utilization strategy. However, these power-related data are stored and owned by different parties. For example, power consumption data are stored in numerous transformer stations across cities; mobility data of the population, which are important indicators of power consumption, are held by mobile companies. Direct data sharing might compromise party benefits, individual privacy and even national security. Inspired by the federated learning scheme from Google AI, we propose a federated learning framework for smart grids, which enables collaborative learning of power consumption patterns without leaking individual power traces. Horizontal federated learning is employed when data are scattered in the sample space; vertical federated learning, on the other hand, is designed for the case with data scattered in the feature space. Case studies show that, with proper encryption schemes such as Paillier encryption, the machine learning models constructed from the proposed framework are lossless, privacy-preserving and effective. Finally, the promising future of federated learning in other facets of the smart grid is discussed, including electric vehicles, distributed generation/consumption and integrated energy systems.

\end{abstract}

}

\maketitle
\IEEEdisplaynontitleabstractindextext
\IEEEpeerreviewmaketitle

\section{Introduction}
\IEEEPARstart{T}{he} 21st century has witnessed an increasingly digitalized and intelligent power grid. The widely distributed sensors, featuring smart meters and global positioning systems, are continuously sending back power-related data from all facets of the grid. Advances in information and communication technologies (ICT) such as 5G have also ensured the timely transmission of these high-resolution data streams back to stakeholders. Big data analytics, attributable to the emergence of such massive data streams, are becoming vastly popular within the past decades. They aim at constructing data-driven models of the physical grid, which help stakeholders establish an informed decision that maximizes their benefits. For example, \cite{Open_Source_Data} recently constructed a comprehensive model of the U.S. power grid, based on the open-source smart grid profiles collected across the Internet, to investigate carbon emissions by 2030. State-of-the-art machine learning models, including deep learning, reinforcement learning and generative networks, are also intensively explored to forecast load and renewable profiles \cite{Load_Forecast}, address grid disturbances \cite{Power_System_Disturbance} and defend against false injection attacks \cite{False_Injection_Attacks}.

One key feature inherent in the big data analytics of smart grids is the natural separation of data by different data-holding parties. This separation in the data space can either be (1) horizontal, where each party holds different training samples but has access to the same set of features, or (2) vertical, where each party shares the same set of training samples but holds different features \cite{Federated_Learning}. Either way, if all the separated data are aggregated into the central server for analysis, the communicational and computational burden would be enormous. The emergence of various distributed learning techniques can mitigate such problems through multi-party collaborated learning, which includes smart task allocation, parallel computing and data sharing mechanisms\cite{Distributed_Learning}. However, most distributed learning frameworks require parties to share their data and models in a direct and unprotected manner (as discussed in \cite{Share_Unprotected}), which leads to the privacy of data being compromised. In a world where data privacy is greatly valued and even legislated (e.g., the General Data Protection Regulation Act in the European Union \cite{GDPR}), it is urgent that all data be secured in the learning process, to minimize data leakage into the hands of eavesdroppers or other parties. 

\begin{figure}[ht]
\centering
\includegraphics[scale=0.3]{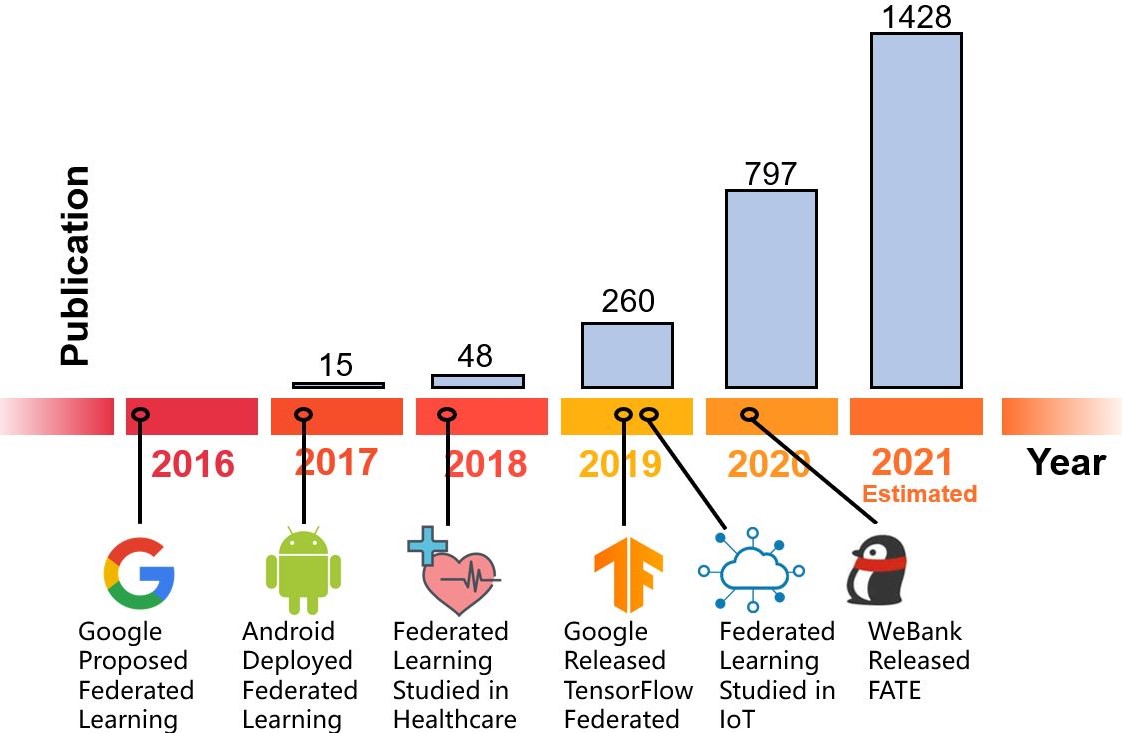}
\caption{Timeline of Federated Learning, which demonstrates important milestones and annual publication statistics.}
\label{Timeline}
\end{figure}

In 2016, Google AI proposed federated learning as a new form of distributed learning, with intentions of learning user behavior across mobile devices \cite{Google_AI}. It echoed the distributed learning scheme in terms of parallel and collaborative data processing, but differed in that only \emph{encrypted} model updates were transferred in each training epoch, while the entire training data remained inside the devices themselves. Both the encryption mechanism and the immobility of user data protected privacy. In the years that follow, various improvements to federated learning emerged, with stronger concerns on addressing the heterogeneity of training data \cite{Non_IID}, computational power \cite{Hetero_Computer_Power} and communicational networks across devices \cite{Hetero_Network}. Yang et al. further extended the concept of federated learning from addressing horizontally-separated to vertically-separated data, and proposed Secure-Boost \cite{Secure_Boost}, a privacy-preserving version of XGBoost \cite{XGBoost}, for nonlinear vertical federated learning. \hyperref[Timeline]{Fig. 1} shows the development of federated learning since the concept was coined. Besides the increasing number of publications by year \cite{WOS}, federated learning has also penetrated into multiple research fields, including smart mobile devices \cite{Android}, healthcare \cite{Healthcare} and Internet of Things \cite{IoT}. Two open-source learning frameworks, TensorFlow Federated \cite{TFF} and FATE \cite{FATE}, were also developed and released in the past 3 years.

Surprisingly, federated learning has rarely been explored in the smart grid domain. The works in \cite{HFL_Load} and \cite{HFL_EV} are among the few implementations of federated learning in electric load forecast and EV-based energy demand prediction, but they merely address the horizontal separation of data, and there is a lack of systematic encryption scheme. The importance of data privacy in the smart grid should be self-evident and significant, as leakage of power traces will lead to grid vulnerabilities being exposed and targeted, affecting the local economy and even national security. Therefore, we hereby propose a federated learning framework for smart grids, to maximally secure power traces in multi-party collaborated learning. The framework is intended to address both vertical and horizontal separations of power traces, and various machine learning models are directly applicable as the built-in learning tool. The framework is demonstrated within a real-life problem setup, where the Zhuhai Power Grid (China) intends to better learn the patterns of power consumption, but the power traces are scattered across inner-city transformer stations, and population mobility data of Zhuhai citizens (which are important indicators of power consumption) are held by commercial mobile companies. Our contributions to the field of smart grid can be summarized as:

\begin{enumerate}[1)]
\item We propose a Horizontal Federated Learning (HFL) sub-framework for multi-area collaborative power consumption forecast, and communication delays are also considered. 

\item We propose two Vertical Federated Learning (VFL) sub-frameworks for collaborative power consumption predictions, based on datasets from two distinct parties. Vertical linear regression is employed when a trusted third-party is needed for augmented data protection, while vertically-federated XGBoost (or SecureBoost) can be employed when better prediction performances are required.

\item The lossless and privacy-preserving properties of both HFL and VFL sub-frameworks are confirmed. 

\end{enumerate}

The incorporation of federated learning into smart grids has profound implications in the energy regime. The concept of the Energy Internet, proposed by Rifkin in 2011 \cite{EI}, has since then brought various energy entities closer than ever, to collectively establish an interconnected and intelligent energy exchange network. The federated learning framework can further help break through the inherent information exchange barriers, allowing for all parties to trustingly collaborate on pattern mining, with an enhanced level of privacy protection.

The rest of the paper is structured as follows. Section II introduces the problem setup of the federated learning framework, which contains both the horizontal and vertical case of data separation. Section III introduces HFL and its applications in the horizontal case. Section IV introduces VFL and its sub-frameworks in the vertical case. Section V concludes the article, and points out the future prospects of federated learning.

\section{Problem Setup}
\begin{figure}
\centering
\includegraphics[scale=0.5]{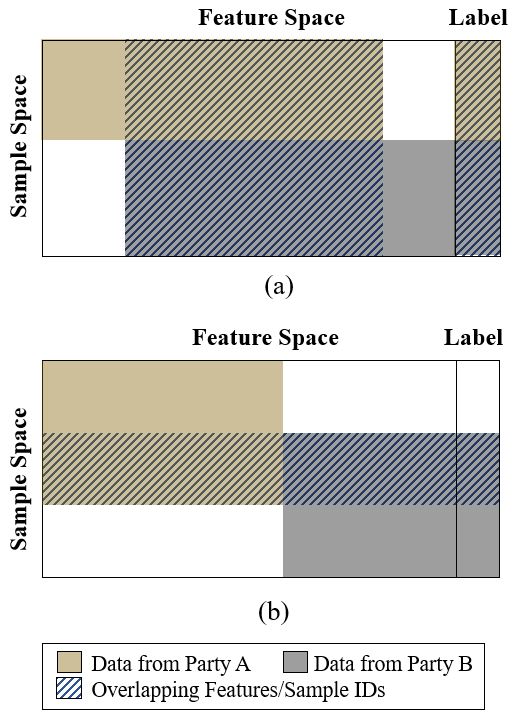}
\caption{Horizontal (a) and vertical (b) separation of data, which respectively share an overlapping set of features and sample IDs.}
\label{Illustration}
\end{figure}

Within the smart grid domain, there are two most commonly observed types of data separation. The first type is known as horizontal data separation, where all parties hold different training samples, but these samples share the same set of features and labels (\hyperref[Illustration]{Fig. 2(a)}). For example, household-level and district-level power consumption data are recorded by numerous sensor networks and transformer stations distributed across cities. Under this construct, all parties are considered homogeneous as they hold similar information of each sample. They can choose to train models separately, but a collaborative training scheme would greatly enrich the training set, contributing to a high generalization performance of the resulting models. The collaboration is easier to achieve since these parties, typically operated by the same power company, have little curious or malicious intentions over each other. Nevertheless, raw data are still preferred not to be transferred directly, as eavesdroppers might easily intercept these information.  

On the other hand, data might be separated in a way that the parties hold different features over the same set of samples (\hyperref[Illustration]{Fig. 2(b)}). This is known as vertical data separation. For example, besides power consumption data, there might also be demographics or weather data that are highly related to power, but held by external parties whose benefits would compromised if data are exchanged. Under this construct, all parties are considered heterogeneous as they hold different feature sets, and collaborative training would be preferred with complementary features as additional inputs. Raw datasets are naturally forbidden to be transferred across parties; even the intermediate model parameters need to be effectively encrypted before transferral.

It can be seen that for both types of data separation, it is necessary to construct a secure and reliable framework for multiple parties to train a model collaboratively. For the ease of discussion, the machine learning goals for each case are concretely formulated as regression problems as follows:

\textbf{Case 1}: For the horizontal separation of data, the constructed model $F^H$ is expected to learn from the feature values of $N$ homogeneous parties $\{X_1,X_2,...,X_N\}$ together with the corresponding labels $\{Y_1,Y_2,...,Y_N\}$, and predict the target values $\mathbf{y}^{H}$ from its feature values $\mathbf{x}$ of the shared feature set:

\begin{equation}
\label{hor}
\setlength\abovedisplayskip{-10pt}
\mathbf{y}^{H} = F_{X_1,...,X_N,Y_1,...,Y_N}^H(\mathbf{x}).
\end{equation}

For each sample, the feature values $\mathbf{x}$ are consistent and transparent to all parties as they share the same feature set. During the training process, raw training data $\{X_1,...,X_N,Y_1,...,Y_N\}$ shall never be transferred across parties. All intermediate results are expected to be properly encrypted before they are transferred.

\textbf{Case 2}: For the vertical separation of data, under a two-party collaboration assumption, the constructed model $F^V$ is expected to learn from the sample-aligned features $X_A$ of Party A and $X_B$ of Party B, together with their labels $Y_B$ (supposedly held by Party B). This model will predict the target values $\mathbf{y}^{V}$, from the feature values of the same sample held by Party A ($\mathbf{x}_{A}$) and Party B ($\mathbf{x}_{B}$):

\begin{equation}
\label{ver}
\setlength\abovedisplayskip{-10pt}
\mathbf{y}^V = F_{X_{A},X_{B},Y_B}^V(\mathbf{x}_{A},\mathbf{x}_{B}).
\end{equation}

During the training process, raw training data $X_{A}$, $X_{B}$, $Y_{B}$, feature values of test samples $\mathbf{x}_{A}$, $\mathbf{x}_{B}$ and feature names shall never be revealed in any way across parties. All intermediate results are expected to be properly encrypted before they are transferred.


In both \textbf{Case 1} and \textbf{Case 2}, mean squared error (MSE) can be employed as a consistent metric for evaluating convergence and testing prediction performances. The proposed framework for training each machine learning model, namely horizontal and vertical federated learning, will ensure a low MSE, and meanwhile minimize data leakage across parties. In Sections III and IV, we will demonstrate principles of the HFL and VFL framework, respectively.

The datasets we use for the case study are retrieved from the Smart City Project of Zhuhai, China (See \hyperref[dataset]{Table I} for details). Among these datasets are the power consumption data from 60 transformer stations distributed across Zhuhai, provided by the Power Supply Bureau with a temporal resolution of 5 minutes. Data of these spatially adjacent transformer stations are natural examples of horizontally separated data (\textbf{Case 1}), and a forecast model can be collaboratively trained due to their similar power consumption patterns. In this case, each training set $X_{i},i \in \{1,2,...,N\}$ in (\ref{hor}) is an $N_S^H \times N_F^H$ matrix, contextualized as the dataset of a transformer station. Here $N_S^H$ is the number of containing samples and $N_F^H$ is the number of shared features. more specifically the historical power consumption records. $Y_{i},i \in \{1,2,...,N\}$ is an $N_S^H \times N_L^H$ matrix, contextualized as the power consumption forecast, where $N_L^H$ is the number of timeslots to be forecasted. $\textbf{x}$, a vector of length $N_F^H$, refers to the historical power consumption of a test sample, and $\textbf{y}^H$ is the corresponding forecast output vector of the model $F^H$ with length $N_L^H$.

Besides power consumption, there is another dataset provided by China Mobile delineating the hourly mobility information of the Zhuhai population. To mitigate ethical concerns, all personal information was erased and instead converted to district-level statistics as features (age group, headcount, etc.) before the dataset was available to us. The mobility dataset can also be temporally aligned with the power consumption dataset with distinct features, which provides a natural example of vertically separated data ($\textbf{Case 2}$). In this case, China Mobile is assigned as Party A, and the Power Supply Bureau, who holds the power consumption data considered as labels, is assigned as Party B. $X_A$ and $X_B$ in (\ref{ver}) are respectively contextualized as feature values for Parties A and B, with shapes $N_S^{V}\times N_{A}^{V}$ and $N_S^{V}\times N_{B}^{V}$. $N_S^{V}$ is the number of common samples; $N_{A}^{V}$ and $N_{B}^{V}$ are the numbers of features held by respectively Parties A and B. $Y_B$ is an $N_S^V \times N_L^V$ matrix, contextualized as the power consumption prediction labels, where $N_L^H$ is the number of labels to be predicted. $\textbf{x}_A$ and $\textbf{x}_B$ are feature vectors of the test sample held by Parties A and B, with lengths $N_{A}^{V}$ and $N_{A}^{V}$ respectively. $\textbf{y}^V$ is the prediction output of model $F^V$ with length $N_L^V$.

Finally, it should be stated that in \textbf{Case 2}, due to a lack of variety in features from the Power Supply Bureau, we scraped weather data in Zhuhai from TimeandDate.com \cite{Time_And_Date} and treated them as if they were proprietary features of $X_{B}$.

\begin{table}[ht]
\label{dataset}
\begin{center}

\caption{Description of the mobility and power consumption dataset used for the problem setup.}
\begin{tabular}{c | c | c }
\hline
\hline
& \textbf{Power Consumption Dataset} & \textbf{Mobility Dataset}\\

\hline
\textbf{Party} & Power Supply Bureau & China Mobile \\
\hline
\textbf{Resolution} & 5 minutes & 1 hour \\
\hline
\textbf{Time Span} & 2019.1 - 2020.10 & 2020.1 - 2020.11 \\
\hline
\textbf{Label} & Power Consumption & ----- \\
\hline
\textbf{Features}* & \makecell{\emph{Temperature}, \emph{Wind speed}, \\ \emph{Humidity}, \emph{Barometer},\\(Historical) Power Consumption} & \makecell{District Headcount, \\ Gender Statistic, \\ Age Statistic (5 bins), \\ Wage Statistic (3 bins)} \\

\hline
\hline 

\end{tabular}
\end{center}
 {\raggedright *The weather-related features (\emph{italicized}) are set as external features of \textbf{Case 2}, in order to enrich the feature set. All mobility data are inferred by individual users' log info in each district. \par}
\end{table}

\section{Horizontal Federated Learning}
In \textbf{Case 1} where data are horizontally separated, it would be ideal if all local datasets are brought together (i.e., $X = X_1 \cup X_2 \cup ... \cup X_N$) for centralized training. However, the strong restraint on raw data exchange poses significant challenges to this training convention. HFL suggests that an alternative learning framework, through distributed local training and \emph{secure} client-server synchronization, is mathematically equivalent to training the combined dataset in a centralized manner. During this process, the raw datasets remain locally stored, and all the synchronized information is safely immune to eavesdroppers.

\subsection{Horizontal Federated Learning Framework}
The client-server based HFL framework \cite{HFL} involves multiple trusted parties training collaboratively as clients, and a central aggregated server for encrypted computation. HFL is built on two trust-related assumptions, both of which are valid in the smart grid context:

\begin{enumerate}
\item All parties are trustworthy to each other. That is, they are not interested in stealing information from each other. In smart grids, all parties holding horizontally separated information are typically regulated by the same state/national grid, and therefore are not viciously competitive towards each other. 

\item The aggregated server is honest-but-curious (or semi-honest). That is, the server will not malignantly hack or contaminate information that is opaque to it; but this server might collect and take advantage of all the ``legal" information for its own benefits. Such an aggregated server is easily available for smart grid data, e.g. many commercial cloud platforms are willing to provide such a service with data securely processed.
 
\end{enumerate}

\begin{figure}[ht]
\centering
\includegraphics[scale=0.3]{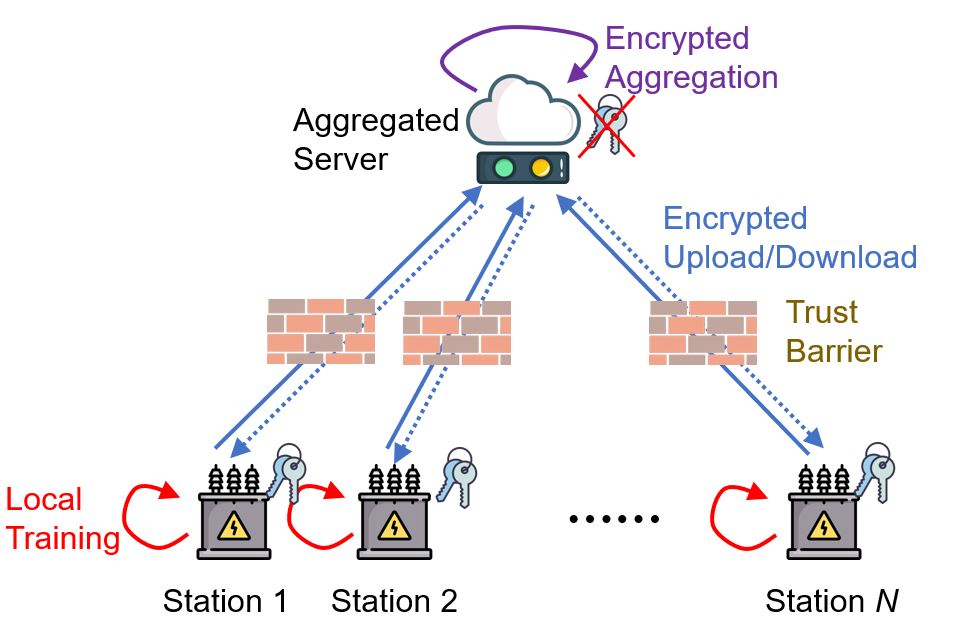}
\caption{Demonstration of collaborative training in horizontal federated learning.}
\label{HFL_Demo}
\end{figure}

An HFL framework can therefore be easily transplanted to the context of smart grids. It trains a unanimous machine learning model $F^H$ in an iterative fashion (See \textbf{Framework 1}), where the updates of parameters are encrypted by, uploaded to, aggregated by and broadcast via a central server in each epoch. Note that all the model parameters are uploaded to and aggregated in the form of ciphertexts, which means that even the server who performs aggregation operations does not have access to the decrypted model updates. After training, each party receives a copy of the final model, as the final product of collaborative learning. Readers can also refer to \hyperref[HFL_Demo]{Fig. 3} for a more intuitive demonstration of the collaborative training process in HFL.

\begin{table}[ht]
\begin{tabular}{l  l }

\hline
\hline
\multicolumn{2}{l}{\textbf{Framework 1: Horizontal Federated Learning}} \\
\hline
\textbf{1} & \makecell[l]{\emph{Aggregated server} distributes the unanimous model $F$ and \\ \quad encrypted parameter initialization $[[\omega_0]]$ to all parties $1,2,...,N$.} \\

\textbf{2} & \textbf{For} each epoch $k=1,...,K$:\\
\textbf{3} & \qquad \textbf{For} \emph{Each Party} $n=1,...,N$:\\
\textbf{4} & \qquad \qquad Decrypt $[[\omega_{k-1}]]$: $\omega_{k-1}=Dec([[\omega_{k-1}]])$\\
\textbf{5} & \qquad \qquad Update $\omega_{k-1}$ by training local dataset $\omega_{k}^n := \omega_{k-1} - \eta g_k^n$\\
\textbf{6} & \qquad \qquad Encrypt $\omega_{k}^n$ into $[[\omega_{k}^n]]$ and send to server.\\ 
\textbf{7} & \qquad \qquad Calculate MSE loss $L_{k}^n$ and send it to server.\\
\textbf{8} & \qquad \textbf{End For} \\
\textbf{9} & \qquad \makecell[l]{\emph{Aggregated server} averages the encrypted parameters \\ \quad (in the cyphertext space) $[[\omega_{k}]]:=\overline{\overline{AVG}}_n[[\omega_{k}^n]]$} \\
\textbf{10} & \qquad \emph{Aggregated server} averages losses $L_{k}:=AVG_n(L_{k}^n)$ \\
\textbf{11} & \qquad \emph{Aggregated server} distributes $L_{k}$ and $[[\omega_{k}]]$ to all parties\\
\textbf{12} & \qquad \textbf{If Agreed on Convergence, Terminate} \\
\textbf{13} & \textbf{End For} \\
\hline
\hline 

\end{tabular}
\end{table}

In \textbf{Framework 1}, $g_k^n$ denotes the calculated gradient in training epoch $k$ for party $n$. $\omega_k$ indicates the averaged model parameters at epoch $k$, which will be the reset parameter values for all parties at epoch $k$. $\omega_{k}^n$ is the local updated parameter for party $n$ at epoch $k$, where $\eta$ is the learning rate. $L_{k}^n$ is the MSE loss for party $n$ at epoch $k$, and $L_{k}$ is its average value across parties.

$[[\cdot]]$ denotes the encryption scheme (consistent throughout this paper), which guarantees security of the HFL framework. The similar update rule of encrypted gradients compared to conventional gradient descent suggests that, the group of real values $\mathbb{R}$ should be homomorphic to the group of ciphertext values $[[\mathbb{R}]]$. The encryption scheme should therefore: (1) allow for all parties to decipher the ciphertexts but not the aggregated server; (2) have bijective mappings between $\mathbb{R}$ and $[[\mathbb{R}]]$; and (3) have well-defined homomorphic operations, at least for additions $\oplus$ and scalar multiplications $\odot$ in the ciphertext space, which is necessary for \emph{server-side} ciphertext averaging $\overline{\overline{AVG}}$:

\begin{equation}
\setlength\abovedisplayskip{-10pt}
Dec([[a]] \oplus [[b]]) = Dec([[a]]) + Dec([[b]])
\end{equation}

\begin{equation}
\setlength\abovedisplayskip{-10pt}
Dec([[a]] \odot n) = Dec([[a \cdot n]])
\end{equation}
where $a$ and $b$ are two independent values in the real space $\mathbb{R}$, and $n$ is a scalar. The Paillier encryption/decryption scheme \cite{Paillier} happens to support such homomorphic operations. Paillier starts by generating a  keypair (public key, private key) based on Number Theory. The public key is available to every entity, which allows all parties to encrypt parameters, and allows the aggregated server to perform ciphertext averaging with homomorphic operations. The private key, necessary for deciphering ciphertexts, is unknown to the server, keeping the parameters unrevealed during the aggregation.

\subsection{Horizontally-Federated Power Consumption Forecasting}
The above HFL framework is implemented in \textbf{Case 1}. We select three transformer stations in Zhuhai, \emph{Leyuan}, \emph{Gaoxin} and \emph{Luoxing}, that are spatially adjacent to each other so that the power consumption patterns are similar.  Historical power consumption (from the past 16 hours) are selected as input features, and the HFL model aims at forecasting the power consumption in the future $1,2,...,7$ hours.

\begin{figure}[ht]
\centering
\includegraphics[scale=0.45]{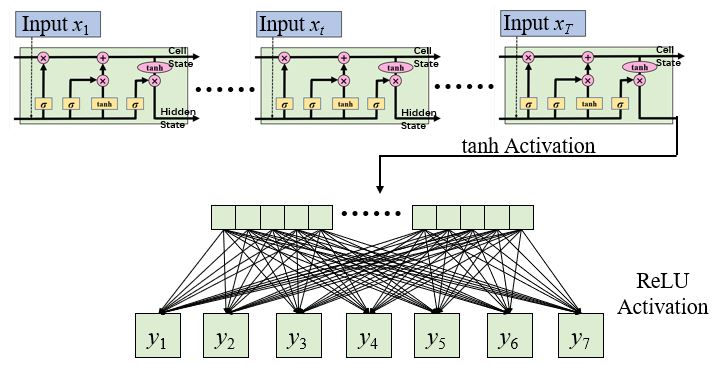}
\caption{LSTM architecture employed in horizontal federated learning.}
\label{LSTM_arc}
\end{figure}

Considering the power of recurrent neural networks in addressing time-series data, we hereby select long-short term memory (LSTM) \cite{LSTM} as the unanimous model $F^H$ to be trained. Details on the LSTM architecture have been elaborated in (\hyperref[LSTM_arc]{Fig. 4}). All time series inputs are passed into identical LSTM units, which encode the inputs into hidden states and cell states using gate logics. The LSTM units are then chained together to yield an intermediate vector, which further passes through activated fully-connected layers to yield the desired output.

\begin{figure}[ht]
\centering
\includegraphics[scale=0.25]{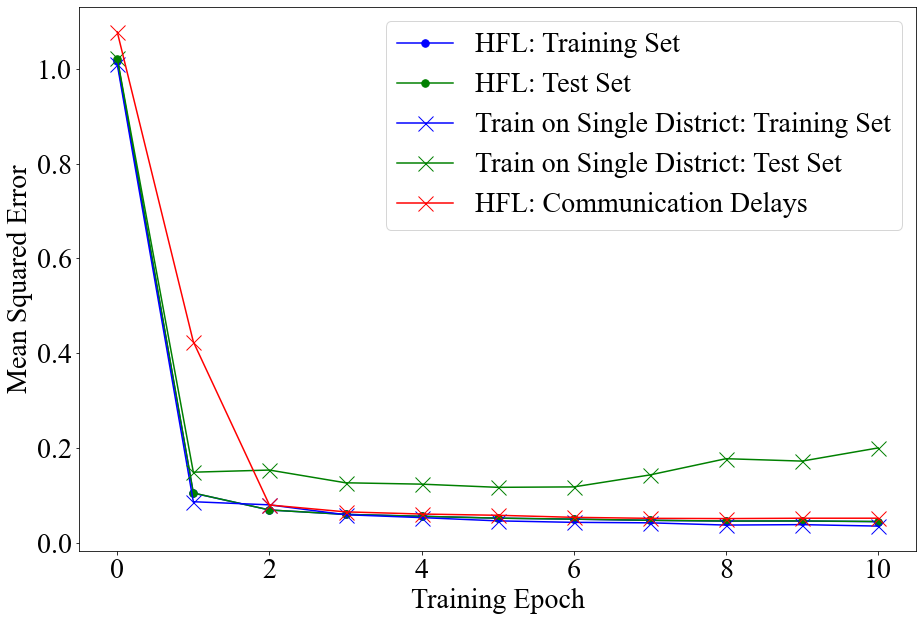}
\caption{LSTM architecture employed in horizontal federated learning.}
\label{HFL_convergence}
\end{figure}

The convergence of HFL with an LSTM architecture has been plotted in \hyperref[HFL_convergence]{Fig. 5}. It can be seen that the HFL framework converges perfectly within 10 epochs with an average MSE of 0.04, and no overfitting is observed. Instead, if each party individually trains its own data, then despite a similarly impressive performance on the training set itself, the as-trained model generalizes poorly to the data of other parties, with a significant level of overfitting on the test set. 

Finally, we consider a particular case where communication delays are present in the data transmission process, which is a common concern in federated learning practices. If in any epoch a party is not able to upload a parameter update in time, the aggregated server discards its updates and instead averages all other parties' updates. Case study shows that when the update failure rate of each party is set to 40\%, this communication delay would result in the MSE error dropping more slowly, but convergence is still reachable within 10 epochs. A slightly higher final MSE level of 0.051 is reached compared the delay-free HFL, but even so the error is still much lower than that of separate training on a single district.

\subsection{Losslessness and Security Analysis}
A federated learning model $F^H$ ($F^V$) is defined to be lossless, if its centralized version $F_0^H$ ( $F_0^V$) would yield the same final model under a similar training setting. This ensures that the model does not ``learn less" than a centralized counterpart. In \textbf{Framework 1}, it is straightforward to deduce the actual parameter update rule from the homomorphic property of Paillier:

\begin{equation}
\setlength\abovedisplayskip{-10pt}
\omega_{k+1} = \omega_{k} - \eta Avg(g_k^n)
\end{equation}
which is identical to that of a centralized gradient descent framework. This proves that $F^H$ and $F_0^H$ would generate the same set of parameters after each epoch training, and therefore confirms HFL's losslessness.

The security of data exchange is guaranteed on the server side as well as in the data transfer process. This is because all information is Paillier-encrypted outside the data-holding parties. However, the  trustworthiness of parties must be guaranteed in HFL, as their uncooperative behavior (say, perform zero update at each epoch in a two-party collaboration) might lead to other parties' private information being exposed.

\section{Vertical Federated Learning}
In \textbf{Case 2} where data are vertically separated, parties would be more reluctant to share raw data and even model parameters, since these parties represent different stakeholder benefits and hold different facets of data. This changes the relationship among parties from being totally trustworthy to honest-but-curious. VFL provides a solution framework where each party keeps private their share of data and model during training, and only necessary results are exchanged with encryption.

\subsection{Vertically-Federated Linear Regression}
Vertically-federated linear regression (VFLR) is one of the plainest attempts at vertical collaboration \cite{VFL_Linear}. It asks for each party to calculate a prediction based on its own set of features, and linearly add these predictions to yield the final estimation:

\begin{equation}
\setlength\abovedisplayskip{-10pt}
\mathbf{y}^V =  \mathbf{x}_{A} \theta^{A} + \mathbf{x}_{B} \theta^{B}.
\end{equation}

The objective function for VFLR to minimize is therefore:

\begin{align}
\label{cites}
\underset{\theta^{A},\theta^{B}}{min} \quad & ||X_{A}\theta^{A}+X_{B}\theta^{B}-y_{B}^{Real}||^2  \notag\\ + & \frac{\lambda}{2}(||\theta_{A}||^2 + ||\theta_{B}||^2)
\end{align}
where $\theta^{A}$ and $\theta^{B}$ are linear coefficients of the mobility features and power consumption features, respectively. $y_{B}^{Real}$ is the actual power consumption label, held by the Power Supply Bureau. $\lambda$ is the regularization parameter. 

The framework of VFLR, following the work of \cite{VFL_Linear}, is introduced in \textbf{Framework 2}. which requires the participation of a totally trustworthy third party. However, we removed the information exchange of the MSE loss, as the convexity of (\ref{cites}) guarantees the convergence and global optimality of this iterative algorithm. The new framework will now be terminated if both parties claim that their trainable parameters have converged.

\begin{table}[ht]
\begin{tabular}{l  l }

\hline
\hline
\multicolumn{2}{l}{\textbf{Framework 2: Vertically-Federated Linear Regression}} \\
\hline
\multicolumn{2}{l}{\emph{A}: China Mobile, \emph{B}: Power Supply Bureau. \emph{C}: Third Party.} \\
\textbf{1} & \makecell[l]{\emph{C} generates a Paillier keypair (public key, private key), \\ \quad and distributes the public key to \emph{A} and \emph{B}.} \\

\textbf{2} & \emph{A} secretly initializes $\theta_0^{A}$. \\
\textbf{3} & \emph{B} secretly initializes $\theta_0^{B}$. \\
\textbf{4} & \textbf{For} each epoch $k=1,2,...,K$: \\
\textbf{5} & \qquad \emph{A} updates and sends $[[X_{A}\theta_k^{A}]]$ to \emph{B}. \\
\textbf{6} & \makecell[l]{\qquad \emph{B} calculates $[[d_k]] = [[X_{A}\theta_k^{A}]] + $ \\ \qquad \quad $[[X_{B}\theta_k^{B}-y_{B}^{Real}]]$ and sends it to \emph{A}.} \\
\textbf{7} & \makecell[l]{\qquad \emph{A} calculates $[[\frac{\partial L}{\partial \theta^{A}}]] = 2 [[d_k]] X_{A} + \lambda[[\theta_k^{A}]]$,\\ \qquad \quad adds mask $[[R_k^{A}]]$ and sends the added value to \emph{C}.} \\ 
\textbf{8} & \makecell[l]{\qquad \emph{B} calculates $[[\frac{\partial L}{\partial \theta^{B}}]] = 2 [[d_k]] X_{B} + \lambda[[\theta_k^{B}]]$,\\ \qquad \quad adds mask $[[R_k^{B}]]$ and sends the added value to \emph{C}.} \\ 
\textbf{9} & \qquad \emph{C} deciphers $\frac{\partial L}{\partial \theta^{A}} + R_{A}$ and sends it to \emph{A}. \\
\textbf{10} & \qquad \emph{C} deciphers $\frac{\partial L}{\partial \theta^{B}} + R_{B}$ and sends it to \emph{B}. \\
\textbf{11} & \qquad \emph{A} secretly updates $\theta_{k+1}^{A}$. \\
\textbf{12} & \qquad \emph{B} secretly updates $\theta_{k+1}^{B}$. \\
\textbf{13} & \qquad \textbf{Terminate if \emph{A} and \emph{B} both agree on convergence.} \\
\textbf{14} & \textbf{End For} \\

\hline
\hline 

\end{tabular}
\end{table}

\begin{figure}[ht]
\centering
\includegraphics[scale=0.3]{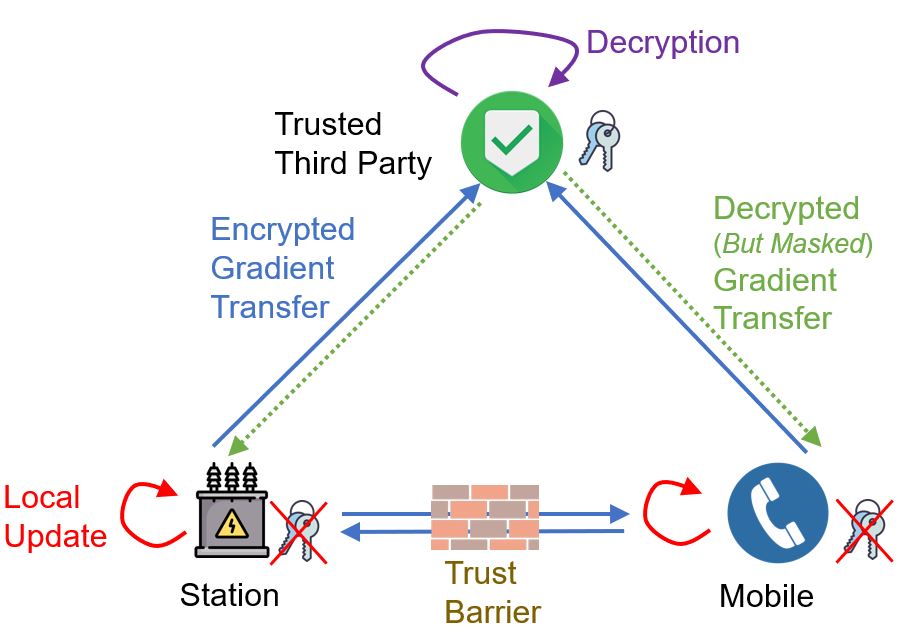}
\caption{Demonstration of collaborative training in vertically-federated linear regression.}
\label{VFLR_Demo}
\end{figure}

During the process, China Mobile and the Power Supply Bureau collaboratively calculate the encrypted intermediate variable $[[d_k]]$, with each party's training data $X_{A}$ (or $X_{B}$), linear parameters $\theta_k^{A}$ (or $\theta_k^{B}$) and public key. The encrypted gradient of MSE loss $[[\frac{\partial L}{\partial \theta^{A}}]]$ (or $[[\frac{\partial L}{\partial \theta^{B}}]]$), due to the homomorphic property of Paillier, happens to both be linearly dependent on $[[d_k]]$, and therefore can be calculated by the corresponding party, without even knowing the decrypted gradient. The encrypted gradient is then sent to the third party for decryption; but a random encrypted mask $[[R_k^{A}]]$ (or $[[R_k^{B}]]$) only known to the corresponding party is added to the gradients, to prevent gradient leakage to the third party. Finally the third party sends back the encrypted values of $\frac{\partial L}{\partial \theta^{A}} + R_{A}$ (or $\frac{\partial L}{\partial \theta^{B}} + R_{B}$), and each party subtracts it by $R_{A}$ (or $R_{B}$) to obtain the real decrypted gradients. Information security of power traces is guaranteed in the sense that no gradient information is obtained by either party (including the third party) until the end of every epoch, where the feature-holding party unmasks the gradients and obtain the final value. Readers can also refer to \hyperref[VFLR_Demo]{Fig. 6} for a more intuitive demonstration of the collaborative training process in VFLR.

\begin{figure}[ht]
\centering
\includegraphics[scale=0.25]{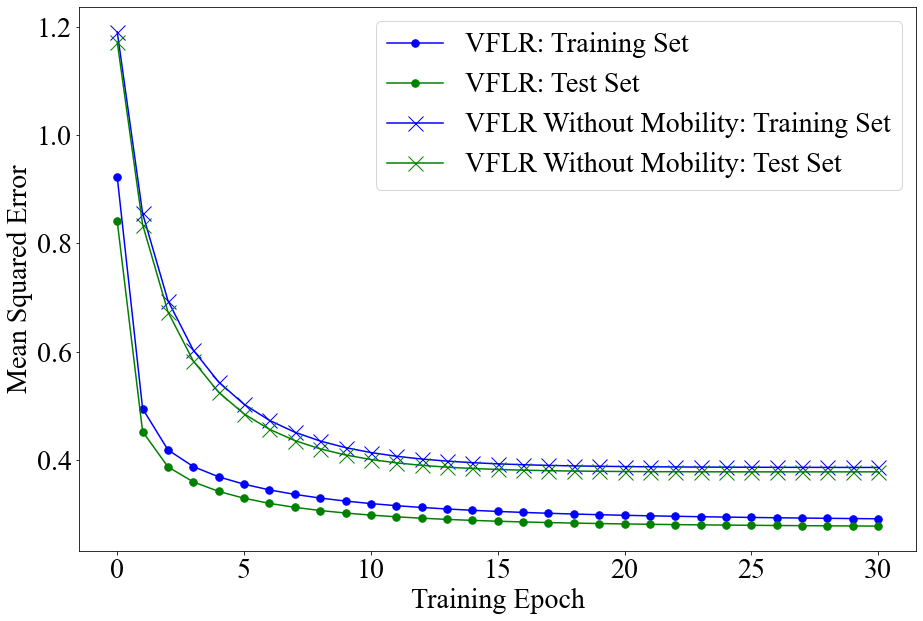}
\caption{Convergence of mean squared error in the training process of vertically-federated linear regression. \emph{MSE values at the 0-th epoch differ with/without the mobility dataset, due to different parameter initializations.}}
\label{VFLR}
\end{figure}

\textbf{Framework 2} is implemented to demonstrate the power consumption prediction performance of VFLR in \textbf{Case 2}. \hyperref[VFLR]{Fig. 7} plots the convergence of VFLR during training, with MSE dropping from 0.92 down to a final 0.28. Validation on the test set also suggests no overfitting issues. Instead, if China Mobile refuses to provide data due to security concerns, the MSE still converges but to a much higher 0.38 on the same training samples.

\subsection{Vertically-Federated XGBoost (SecureBoost)}
One apparent drawback of VFLR is its deficiency in learning the nonlinearity of data, which can be addressed by the introduction of regression trees. Regression trees categorize data from ``root node" down to ``leaf nodes" by their feature values, where each leaf node corresponds to a prediction value. XGBoost is one of the most successful innovations in regression tree, which consists of $N$ separate regression trees $\{F_1^V,F_2^V,...,F_n^V\}$, each predicting the residual value of all previous trees \cite{XGBoost}:

\begin{align}
\setlength\abovedisplayskip{-10pt}
\textbf{y}^V = &F_1^V(\textbf{x}_{A},\textbf{x}_{B}) + F_2^V(\textbf{x}_{A},\textbf{x}_{B}) + ... \notag \\
& + F_n^V(\textbf{x}_{A},\textbf{x}_{B}), n=1,2,...,N. 
\end{align}

Another unique innovation in XGBoost lies in the split-finding process, which 
determines how data are further categorized at each node. Instead of searching the sample space individually for a greedy separation, XGBoost pre-divides training data into bins for each feature, and decides which edge of bins yields the maximum information gain. First-order gradients $g_i$ and second-order gradients $h_i$ of the current MSE losses are also aggregated by the bins:

\begin{equation}
\setlength\abovedisplayskip{-10pt}
G_{vb} = \Sigma_{\{i|s_{v,b-1}<x_{i,v} \leq s_{v,b}\}}g_i, \forall v,b
\end{equation}

\begin{equation}
\setlength\abovedisplayskip{-10pt}
H_{vb} = \Sigma_{\{i|s_{v,b-1}<x_{i,v} \leq s_{v,b}\}}h_i, \forall v,b
\end{equation}
where $x_{i,v}$ is the value of the $v$-th feature of the $i$-th sample. $(s_{v,b-1},s_{v,b}]$ is the $b$-th bin of the $v$-th feature.

XGBoost is the perfect candidate for nonlinear learning of vertically separated data, as training data $X_{A}$ and $X_{B}$ are not heavily interwined by features as one would expect: each node-splitting operation involves only a split of one feature, and hence requires only one party to decide the property of the left/right child node. Split-finding, the core of XGBoost training, is extended to the vertically-federated version (or SecureBoost \cite{Secure_Boost}) shown in \textbf{Framework 3}. Note that different than VFLR, a third party is not required for encryption and decryption. Instead, the Power Supply Bureau holds the private key and dominates the split finding process.

\begin{table}[ht]
\begin{tabular}{l  l }

\hline
\hline
\multicolumn{2}{l}{\textbf{Framework 3: Split Finding in SecureBoost}} \\
\hline
\multicolumn{2}{l}{\emph{A}: China Mobile, \emph{B}: Power Supply Bureau.}\\
\textbf{1} & \makecell[l]{\emph{B} generates a Paillier keypair (public key, private key), \\ \quad and distributes the public key to \emph{A}.} \\
\textbf{2} & \emph{A} and \emph{B} identify all training samples $i$ that belong to this node. \\
\textbf{3} & \emph{B} calculates $g_i$ and $h_i$. \\
\textbf{4} & Each party calculates $G_{vb}$ and $H_{vb}$ for features $v$ they hold. \\
\textbf{5} & \emph{A} distributes $[[G_{vb}]]$ to \emph{B}. \\
\textbf{6} & \emph{B} decrypts $[[G_{vb}]]$ from \emph{A} with the private key. \\
\textbf{7} & (\emph{B} performs the rest): \\
\textbf{8} & \textbf{For} each \emph{feature index} $v=1,2,...,V_{A}+V_{B}$: \\
\textbf{9} & \qquad $g:=\Sigma g_i,h:=\Sigma h_i,g_l:=0,h_l:=0,g_r:=g,h_r:=h$. \\
\textbf{10} & \qquad \textbf{For} each \emph{splitting edge index} $b=1,2,...,B_v$:\\
\textbf{11} & \qquad \qquad $g_l := g_l + G_{vb},g_r := g_r - G_{vb}$\\
\textbf{12} & \qquad \qquad $Score = \mathbf{max} \{Score,\frac{g_l^2}{h_l+\lambda}+\frac{g_r^2}{h_r+\lambda}-\frac{g^2}{h +\lambda}\}$\\
\textbf{13} & \qquad \textbf{End For}\\
\textbf{14} & \textbf{End For} \\
\textbf{15} & \makecell[l]{Record the best \emph{feature index} and \emph{splitting edge index}. If this feature \\ \quad belongs to \emph{A}, inform \emph{A} of these information.}\\
\hline
\hline 

\end{tabular}
\end{table}

For each node in the SecureBoost Tree that needs to be split upon, China Mobile and the Power Supply Bureau first collaboratively identify the set of training samples that fall onto this node. Then, each party calculates its bin statistics $G_{vb}$ and $H_{vb}$ for all features $v$ they hold, which are aggregated to the Power Supply Bureau. Every bin edge for every feature represents a potential node split, and the $Score$ metric, an indicator of splitting-induced information gain, is tracked to find the splitting strategy among all splitting candidates. In the calculation of $Score$, the set of $G_{vb}$'s ($H_{vb}$'s) are further aggregated into the left-child statistic $g_l$ ($h_l$) and right-child statistic $g_r$ ($h_r$) generated from the splitting.

\begin{figure}[ht]
\centering
\includegraphics[scale=0.4]{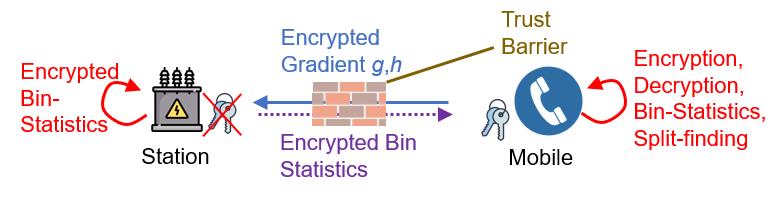}
\caption{Demonstration of collaborative training in SecureBoost.}
\label{SecureBoost_Demo}
\end{figure}

Despite that the Power Supply Bureau is more dominant in the training process (See \hyperref[SecureBoost_Demo]{Fig. 8} for demonstration), it is opaque to the actual interpretation of the selected feature index and splitting edge. This guarantees that information exchanged between parties cannot be traced back to substantive conclusions of the opposite party's data.

\begin{figure}[ht]
\centering
\includegraphics[scale=0.25]{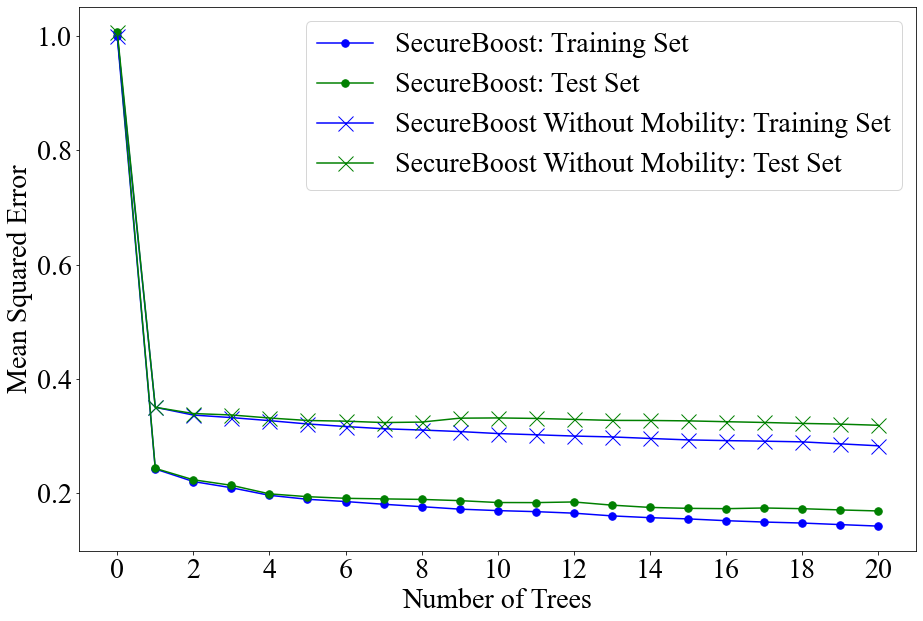}
\caption{Convergence of mean squared error in the training process of SecureBoost.}
\label{SB}
\end{figure}

Similar to VFLR, SecureBoost is also implemented in \textbf{Case 2} to confirm its effectiveness, where the convergence of MSE is plotted in \hyperref[SB]{Fig. 9}. It can be seen that with SecureBoost, MSE loss has decreased all the way down to 0.14 with the growing number of trees $N$ in the model, which is a 50\% improvement compared to VFLR. Furthermore, the majority of MSE drop takes place right after training the first tree, by which time the MSE loss is already smaller than that of the final VFLR model. This indicates that we only need to train the first tree if computational efficiency is required. Finally similar to VFLR, if China Mobile feels its data might be compromised and withdrew from the collaboration, then the MSE loss will converge to a much higher 0.28.

\begin{figure}[ht]
\centering
\includegraphics[scale=0.5]{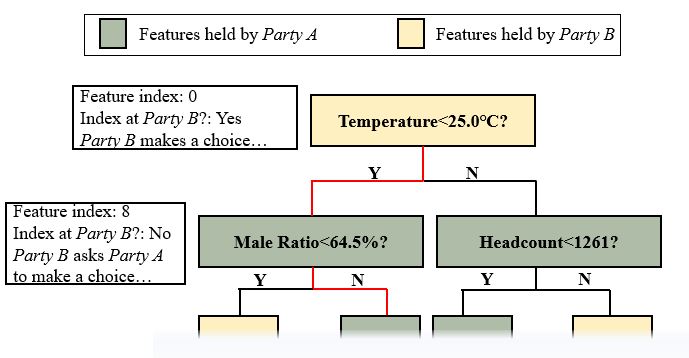}
\caption{Collaborative prediction process with SecureBoost.}
\label{SB_pred}
\end{figure}

After training, the Power Supply Bureau holds the entire XGBoost model, except on nodes when the feature value of a split is held by China Mobile, in which case it only has the index of feature and splitting edges. Hence, it still requires collaborative efforts to predict the labels for unknown test samples. \hyperref[SB_pred]{Fig. 10} demonstrates the prediction process with the as-trained model of \textbf{Case 2}. When a test sample falls onto a non-leaf node, the Power Supply Bureau first locates the feature index on which to split the node. If the feature index belongs to Power Supply Bureau already, it then looks up the threshold value to decide which child node the test sample belongs to. If instead the feature index falls onto China Mobile, then Power Supply Bureau would request China Mobile to check out the test sample and inform whether it should belong to the left or right child node. Finally the test sample would reach onto a leaf node, which contains the prediction value of this sample on this particular tree.

\subsection{Losslessness and Security Analysis}
Similar to the HFL framework, VFLR and SecureBoost are identical to their centralized version in terms of prediction power. For VFLR, it is straightforward to deduce that the decrypted gradient

\begin{equation}
\begin{split}
& Dec([[\frac{\partial L}{\partial \theta^{A}}]]) = 2d_k X_{A} + \lambda \theta_k^{A} \\
& = 2(X_{A}\theta_k^{A}+X_{B}\theta_k^{B}-y_{B}^{Real})X_{A} + \lambda \theta_k^{A}
\end{split}
\end{equation}
is exactly the derivative of (\ref{cites}) over $\theta_{A}$ at point $\theta_k^{A}$. For SecureBoost, as the Power Supply Bureau acquires information on 
the bin statistics $G_{vb}$,$H_{vb}$ after decryption, the training steps are exactly the same compared to a centralized counterpart. The above analysis ensures the losslessness of both VFL frameworks.

The VFLR framework, due to the introduction of a totally trustworthy third party to distribute the public keys, can be considered secure enough. Neither party can decrypt the gradients until the very last step of iteration; and even the third party, who is responsible for decrypting the gradients, will be blocked from knowing the gradients due to the existence of random masks.

For SecureBoost, China Mobile (or Party \emph{A}) is opaque to private information of the model, as it does not even have access to the leaf nodes, which are fundamental to the prediction values, in the first place. The Power Supply Bureau, despite knowing every feature index for splitting in the regression tree, cannot infer the representations of features nor the exact bin edges. Besides, regression problems hardly cause information leakage compared to classification problems, where the binary classification of nodes alone provides minimal insights on the private properties of data \cite{Secure_Boost}.

Finally, unlike HFL, VFL does not highlight the importance of addressing communication delays. This is because in VFL there are much fewer parties, each of which holds vital parameters not publicly revealed. Hence, a training epoch is not considered complete until exchange information from all parties has successfully arrived at the destination party.

\section{Conclusion}
In this paper, we proposed a systematic federated learning framework, that addresses both the horizontal and vertical separation of data in a distributed fashion. To protect user privacy and secure power traces, raw data are always stored locally to prevent data leakage, while the model parameters and sample statistics are Paillier-encrypted before the exchange. Two case studies centered around the real-life problem setup confirms the effectiveness of the proposed federated learning sub-frameworks, whose lossless property is also validated through analysis. 

Although this paper merely studies the case of federated learning in power consumption prediction, this framework in fact has further implications in almost every facet of smart grids. This is especially true with the emergence of the concept ``Energy Internet", which aims to interconnect all energy components. Below we have illustrated 3 potential applications of this framework in smart grids:

\begin{enumerate}[1)]
\item Owners of \emph{electric vehicles} are reluctant to reveal their trajectories or battery conditions frequently to the public. However, understanding the data of EVs is critical to the optimal siting of charging stations and the design of smart navigation systems. Federated learning helps to perform local data mining of these homogeneous vehicles, while sharing the model updates on each device.

\item With the prevalence of distributed power systems, owners of \emph{distributed generators/consumers} may not willingly provide their electricity inputs/outputs, but these data would be critical to the safe and economic operation of smart grids. Federated learning would reassure the distributed participants in collaborative pattern mining, without exposing raw data to each other.

\item In the \emph{integrated energy system}, electricity is dynamically coupled with other energy systems (heat, cool, gas, etc.) Different energy systems implicitly refer to different stakeholders, and their benefits will hopefully be secured if federated learning is implemented to secure the information exchange.

\end{enumerate}

Federated learning can further extend its power by leveraging state-of-the-art cloud-computing and edge-computing technologies. We will explore these topics in the future.


\begin{thebibliography}{1}

\bibitem{Open_Source_Data}
``Modeling a Clean Energy Future for the United States", Breakthrough Energy, 2021 [Online]. Available: \href{https://science.breakthroughenergy.org/}{https://science.breakthroughenergy.org/}.

\bibitem{Load_Forecast}
Y. Chen, Y. Wang, D. Kirschen and B. Zhang, ``Model-Free Renewable Scenario Generation Using Generative Adversarial Networks," 2019 IEEE Power \& Energy Society General Meeting (PESGM), Atlanta, GA, USA, 2019, pp. 1-1.

\bibitem{Power_System_Disturbance}
S. Wang and H. Chen, ``A novel deep learning method for the classification of power quality disturbances using deep convolutional neural network," in Applied Energy, vol. 235, pp. 1126-1140, Feb 2019.

\bibitem{False_Injection_Attacks}
Y. Chen, Y. Wang, D. Kirschen and B. Zhang, ``Model-Free Renewable Scenario Generation Using Generative Adversarial Networks," in IEEE Transactions on Power Systems, vol. 33, no. 3, pp. 3265-3275, May 2018.

\bibitem{Federated_Learning}
Q. Yang, Y. Liu, Y. Cheng, Y. Kang, T. Chen and H. Yu, ``Federated Learning," in Synthesis Lectures on Artificial Intelligence and Machine Learning, vol. 13, no. 3, pp. 1-207, 2019.

\bibitem{Distributed_Learning}
D. Peteiro-Barral and B. Guijarro-Berdinas, ``A survey of methods for distributed machine learning," in Progress in Artificial Intelligence, vol. 2, no. 1, pp. 1-11, 2013.

\bibitem{Share_Unprotected}
P. Vepakomma, O. Gupta, A. Dubey and R. Raskar, ``Reducing leakage in distributed deep learning for sensitive health data," arXiv preprint arXiv:1812.00564, 2019.

\bibitem{GDPR}
P. Voigt and V. dem Bussche, ``The EU General Data Protection Regulation (GDPR)," 1st Ed., Cham: Springer International Publishing, vol. 10, pp. 3152676, 2017.

\bibitem{Google_AI}
J. Konecny, H. B. McMahan, F. X. Yu, A. T. Suresh and D. Bacon, ``Federated Learning: Strategies for Improving Communication Efficiency," arXiv preprint arXiv:1610.05492, 2016.

\bibitem{Non_IID}
F. Sattler, S. Wiedemann, K. -R. Müller and W. Samek, ``Robust and Communication-Efficient Federated Learning From Non-i.i.d. Data," in IEEE Transactions on Neural Networks and Learning Systems, vol. 31, no. 9, pp. 3400-3413, Sept. 2020.

\bibitem{Hetero_Computer_Power}
Y. Zhan, P. Li and S. Guo, ``Experience-Driven Computational Resource Allocation of Federated Learning by Deep Reinforcement Learning," 2020 IEEE International Parallel and Distributed Processing Symposium (IPDPS), New Orleans, LA, USA, 2020, pp. 234-243.

\bibitem{Hetero_Network}
M. S. H. Abad, E. Ozfatura, D. GUndUz and O. Ercetin, ``Hierarchical Federated Learning ACROSS Heterogeneous Cellular Networks," ICASSP 2020 - 2020 IEEE International Conference on Acoustics, Speech and Signal Processing (ICASSP), Barcelona, Spain, 2020, pp. 8866-8870.


\bibitem{Secure_Boost}
K. Cheng, T. Fan, Y. Jin, Y. Liu, T. Chen and Q. Yang, ``SecureBoost: A Lossless Federated Learning Framework," arXiv preprint arXiv:1901.08755, 2019.

\bibitem{XGBoost}
T. Chen and C. Guestrin, ``XGBoost: A Scalable Tree Boosting System," KDD '16: Proceedings of the 22nd ACM SIGKDD International Conference on Knowledge Discovery and Data Mining, New York, NY, USA, pp. 785–794, 2016.

\bibitem{WOS}
Web of Science [Online]. Available: \href{https://apps.webofknowledge.com/}{https://apps.webofknowledge.com/}. Accessed Mar 31, 2021.

\bibitem{Android}
B. McMahan and D. Ramage, ``Federated Learning: Collaborative Machine Learning without Centralized Training Data", Google AI Blog, 2017 [Online]. Available: \href{https://ai.googleblog.com/2017/04/federated-learning-collaborative.html}{https://ai.googleblog.com/2017/04/federated-learning-collaborative.html}

\bibitem{Healthcare}
T. S. Brisimi, R. Chen, T. Mela et al, ``Federated learning of predictive models from federated Electronic Health Records," in International Journal of Medical Informatics, vol. 112, pp. 59-67, 2018.

\bibitem{IoT}
Y. Lu, X. Huang, Y. Dai, S. Maharjan and Y. Zhang, ``Blockchain and federated learning for privacy-preserved data sharing in industrial IoT," in IEEE Transactions on Industrial Informatics, vol. 16, no. 6, pp. 4177-4186, 2019.

\bibitem{TFF}
``Tensorflow Federated (TFF): Machine Learning on decentralized data," Google, 2019 [Online]. Available: \href{https://www.tensorflow.org/federated}{https://www.tensorflow.org/federated}.

\bibitem{FATE}
``Federated AI Technology Enabler (FATE)," WeBank AI Department, 2020 [Online]. Available: \href{https://github.com/FederatedAI/FATE}{https://github.com/FederatedAI/FATE}.

\bibitem{HFL_Load}
A. Taïk and S. Cherkaoui, ``Electrical Load Forecasting Using Edge Computing and Federated Learning," ICC 2020 - 2020 IEEE International Conference on Communications (ICC), Dublin, Ireland, 2020, pp. 1-6.

\bibitem{HFL_EV}
Y. M. Saputra, D. T. Hoang, D. N. Nguyen, E. Dutkiewicz, M. D. Mueck and S. Srikanteswara, ``Energy Demand Prediction with Federated Learning for Electric Vehicle Networks," 2019 IEEE Global Communications Conference (GLOBECOM), Waikoloa, HI, USA, 2019, pp. 1-6.

\bibitem{EI}
J. Rifkin, ``The third industrial revolution: how lateral power is transforming energy, the economy, and the world," in Macmillan, 2011.

\bibitem{Time_And_Date}
``World Temperatures — Weather Around The World," [Online]. Available: \href{https://www.timeanddate.com/weather/}{https://www.timeanddate.com/weather/}.

\bibitem{HFL}
H. B. McMahan, E. Moore, D. Ramage et al, ``Communication-efficient learning of deep networks from decetralized data,"  arXiv preprint arXiv:1602.05629, 2016.

\bibitem{Paillier}
P. Paillier, ``Public-key cryptosystems based on composite degree residuosity classe," Proc. of International Conference on the Theory and Applications of Cryptographic Techniques, pp. 223-238, 1999.

\bibitem{LSTM}
S. Hochreiter and J. Schmidhuber, ``Long short-term memory," in Neural Computation, vol. 9, no. 8, pp. 1735–1780, 1997. 

\bibitem{VFL_Linear}
Q. Yang, Y. Liu, T. Chen and Y. Tong, ``Federated Machine Learning: Concept and Applications," in ACM Transactions on Intelligent Systems and Technology, vol. 10, no. 2, 2019.

\end{thebibliography}
\end{document}